\pdfoutput=1

\documentclass[11pt]{article}

\usepackage[preprint]{acl}

\usepackage{times}
\usepackage{latexsym}
\usepackage{siunitx}
\usepackage[T1]{fontenc}

\usepackage[utf8]{inputenc}

\usepackage{microtype}

\usepackage{inconsolata}

\usepackage{graphicx}

\usepackage{amsmath}
\usepackage{bm}
\usepackage{xspace}
\usepackage{xstring}
\usepackage{booktabs}
\usepackage{subcaption}
\usepackage{multirow}
\usepackage{soul}
\usepackage{hyperref}
\usepackage{amssymb}
\usepackage[normalem]{ulem}
\useunder{\uline}{\ul}{}

\title{Multilingual Pretraining for Pixel Language Models}

\author{Ilker Kesen$^{\dagger}$ \ \ Jonas F. Lotz$^{\dagger,\ddagger}$ \ \ Ingo Ziegler$^{\dagger}$ \ \ Phillip Rust$^{\dagger}$ \ \ Desmond Elliott$^{\dagger}$\\
$^{\dagger}$Department of Computer Science, University of Copenhagen \\
$^{\ddagger}$ROCKWOOL Foundation Research Unit \\
\texttt{ilke@di.ku.dk}}

\newcommand{\scdigits}[1]{\scalebox{0.8}{#1}}

\newcommand{\mpixel}{\textsc{pixel-m\scdigits{4}}\xspace}
\newcommand{\pixel}{\textsc{pixel}\xspace}
\newcommand{\pixelbigrams}{\textsc{pixel-bigrams}\xspace}
\newcommand{\pixelbgs}{\textsc{pixel-bigrams}\xspace}

\newcommand{\ara}{\textsc{ara}\xspace}
\newcommand{\arz}{\textsc{arz}\xspace}
\newcommand{\eng}{\textsc{eng}\xspace}
\newcommand{\cop}{\textsc{cop}\xspace}
\newcommand{\jpn}{\textsc{jpn}\xspace}
\newcommand{\hin}{\textsc{hin}\xspace}
\newcommand{\kor}{\textsc{kor}\xspace}
\newcommand{\vie}{\textsc{vie}\xspace}
\newcommand{\zho}{\textsc{zho}\xspace}
\newcommand{\tam}{\textsc{tam}\xspace}

\newcommand{\fra}{\textsc{fra}\xspace}

\newcommand{\deu}{\textsc{deu}\xspace}
\newcommand{\bul}{\textsc{bul}\xspace}

\newcommand{\rus}{\textsc{rus}\xspace}

\newcommand{\tur}{\textsc{tur}\xspace}
\newcommand{\urd}{\textsc{urd}\xspace}
\newcommand{\ben}{\textsc{ben}\xspace}
\newcommand{\fin}{\textsc{fin}\xspace}

\newcommand{\tel}{\textsc{tel}\xspace}
\newcommand{\srp}{\textsc{srp}\xspace}
\newcommand{\ukr}{\textsc{ukr}\xspace}
\newcommand{\uig}{\textsc{uig}\xspace}
\newcommand{\uzn}{\textsc{uzn}\xspace}
\newcommand{\kir}{\textsc{kir}\xspace}

\newcommand{\greek}{\textsc{ell}\xspace}
\newcommand{\bod}{\textsc{bod}}
\newcommand{\hye}{\textsc{hye}}
\newcommand{\heb}{\textsc{heb}}

\newcommand{\bert}{\textsc{bert}\xspace}
\newcommand{\monobert}{\textsc{bert-mono}\xspace}
\newcommand{\canine}{\textsc{canine-s}\xspace}

\newcommand{\fngithub}[1]{%
  \StrBehind{#1}{https://github.com/}[\githubPathTemp]%
  \mbox{%
    \xspace
    \raisebox{-0.1ex}{\includegraphics[width=0.8em, height=0.8em]{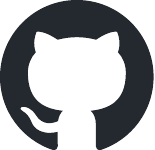}}%
    \hspace{0.2em}%
    \href{#1}{\footnotesize\texttt{\nolinkurl{\githubPathTemp}}}%
  }%
}

\begin{document}
\maketitle
\begin{abstract}

Pixel language models operate directly on images of rendered text, eliminating the need for a fixed vocabulary.
While these models have demonstrated strong capabilities for downstream cross-lingual transfer, multilingual pretraining remains underexplored.
We introduce \mpixel, a model pretrained on four visually and linguistically diverse languages: English, Hindi, Ukrainian, and Simplified Chinese.
Multilingual evaluations on semantic and syntactic tasks show that \mpixel outperforms an English-only counterpart on non-Latin scripts.
Word-level probing analyses confirm that \mpixel captures rich linguistic features, even in languages not seen during pretraining.
Furthermore, an analysis of its hidden representations shows that multilingual pretraining yields a semantic embedding space closely aligned across the languages used for pretraining.
This work demonstrates that multilingual pretraining substantially enhances the capability of pixel language models to effectively support a diverse set of languages.

\end{abstract}

\section{Introduction}
\label{sec:introduction}
\begin{figure}[!ht]
  \centering
  \includegraphics[width=\columnwidth]{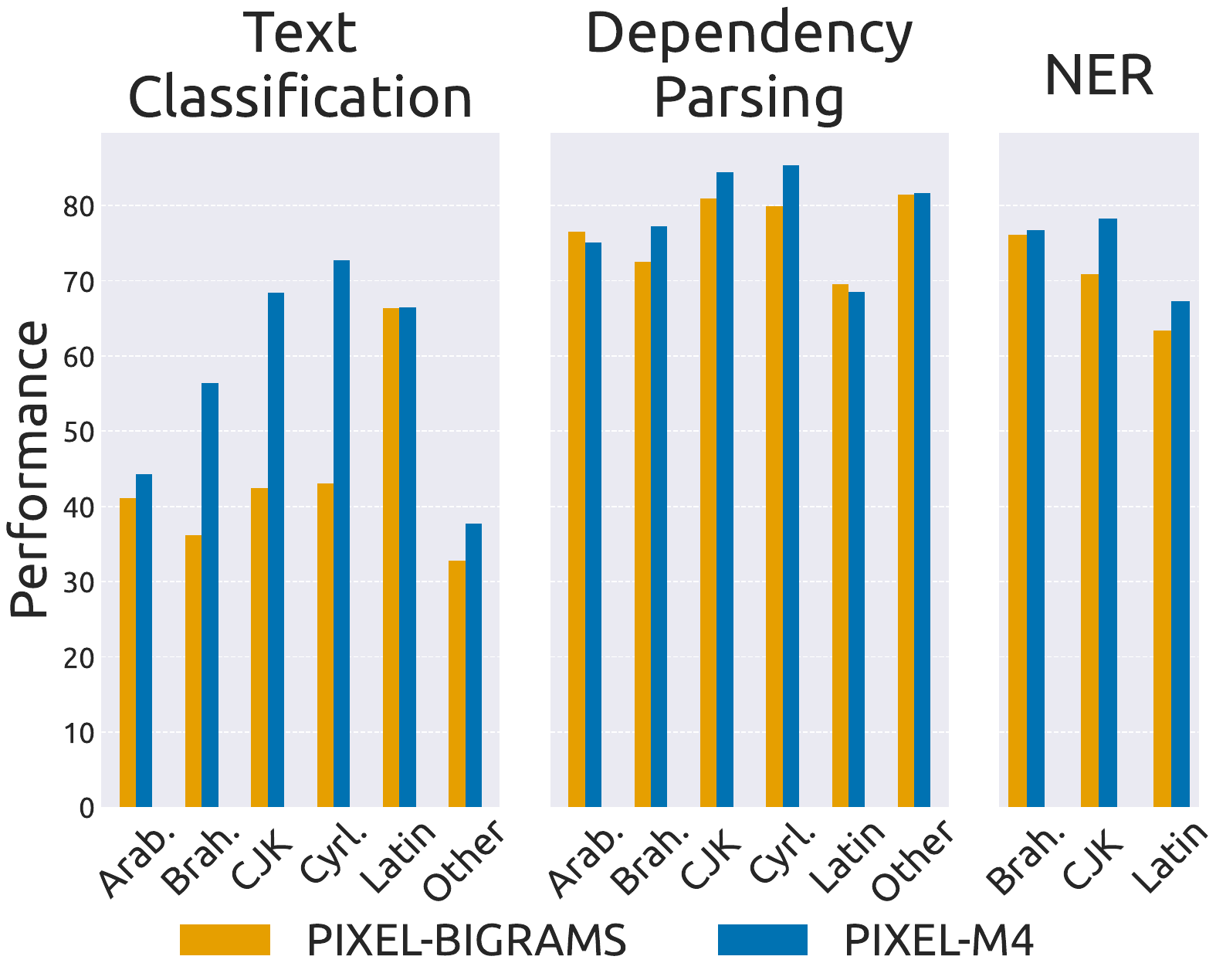}
  \caption{
    Average performance across tasks comparing \mpixel and \pixelbigrams grouped by scripts: Arabic, Brahmic, Chinese-Japanese-Korean, Cyrillic, Latin, and others.
    Both models share the same architecture and hyperparameters, but \mpixel is pretrained in four visually and linguistically diverse languages: English, Hindi, Ukrainian and Simplified Chinese. \mpixel performs better in almost all non-Latin script languages without sacrificing Latin-script performance.
    }
  \label{fig:fig1}
\end{figure}

Visually-rendered text has emerged as an alternative to sub-word tokenization for language models~\citep{salesky-etal-2021-robust,rust2023pixel}.
In comparison to sub-word tokenization, processing visually-rendered text enables models to transfer to unseen languages without needing to initialize new embeddings~\cite{dobler-de-melo-2023-focus}, or relying on back-off mechanisms based on  bytes~\cite{xue-etal-2022-byt5} or characters~\cite{clark-etal-2022-canine}.    
Previous work on pixel-based language models has predominantly focused on monolingual pretraining on English data~\cite{rust2023pixel,lotz-etal-2023-text}, with related efforts extending to multilingual pretraining for machine translation~\cite{salesky-etal-2023-multilingual}. 
Given evidence that pixel-based models facilitate positive transfer through visual similarity \citep{lotz-etal-2025-overcoming, munoz-ortiz-etal-2025-evaluating}, we investigate multilingual pretraining for general-purpose representation learning specifically by selecting only one language per script. This approach is particularly valuable for low-resource languages that can benefit from transfer via visually similar, high-resource languages.

We present \mpixel: a multilingual version of \pixel \citep{rust2023pixel}.
\mpixel is pretrained on four
equally-sized amounts of 
visually diverse scripts sourced from mC4~\cite{xue-etal-2021-mt5}: English (Latin script), Hindi (Devanagari script), Simplified Chinese (Han script), and Ukrainian (Cyrillic script).
These scripts were chosen to represent abugida, alphabetic, logographic/logosyllabic writing systems, covering billions of speakers.
Furthermore, not only do these scripts represent visual diversity, they also represent grammatical diversity, covering Balto-Slavic, Indo-Iranian, Germanic, and Sino-Tibetan languages.

In downstream task experiments, we investigate the ability of \mpixel to transfer to new languages in three conditions (i) same-script; (ii) related-script; and (iii) unrelated scripts to better understand what is gained by multilingual pretraining.\footnote{The downstream task languages also cut across different language families, e.g. Indo-European, Sino-Tibetan, and Turkic. However, we focus on script transfer, given the visual nature of the data processed by \mpixel.}
The same-script experiments focus on
Simplified Chinese (Han), Hindi (Devanagari), and various Latin and Cyrillic script languages.
The related-script experiments include Japanese
and Brahmic script languages;
while the unrelated-script experiments focus on Armenian, Greek, Korean and languages using the abjad writing system (e.g. Arabic and Hebrew).
Compared to its monolingually-pretrained equivalent, \pixelbigrams \citep{lotz-etal-2023-text}, we find consistent improvements in performance for almost all non-Latin script languages on 
text classification, dependency parsing and named entity recognition. 

We conduct word-level probing experiments using \textsc{linspector}~\cite{csahin-etal-2020-linspector} to compare differences in linguistic understanding across 15 languages from multilingual versus monolingual pretraining. 
We find that \mpixel captures linguistic features more effectively than \pixelbigrams, both for seen scripts (e.g., Russian and Macedonian) and unseen scripts (Arabic, Armenian, Greek).
Additionally, an exploration of \mpixel's embedding space reveals
that earlier layers primarily encode visual information,
while deeper layers shift toward semantic understanding,
corroborating earlier observations by \citet{tatariya-etal-2024-pixology}.
Through cross-lingual retrieval experiments, we find that \mpixel has learned a semantic representation space that is shared across the pretraining languages.

In short, the main contributions of this paper are:
\begin{itemize}
  \item We present the first multilingually-pretrained general-purpose pixel language model,\footnote{Code and models: \fngithub{https://github.com/ilkerkesen/pixel-m4}}
  trained over four visually and linguistically diverse languages.
  \item Experiments on syntactic and semantic tasks show consistent improvements for non-Latin script languages compared to previous \pixel language models.
  \item Word-level probing analyses show that multilingual pretraining produces representations that capture more linguistic features across languages, such as case marking, part-of-speech tags, and verb tense.
  \item Sentence-level analyses of the learned hidden representations
  reveal that \mpixel has learned a representation space highly aligned between a subset of its pretraining languages.
\end{itemize}

\section{\mpixel}
\label{sec:method}

\subsection{Pretraining Data}
\label{sec:method:data}
Following our motivation to explore multilingual pretraining through a diverse selection of scripts rather than a large range of languages, \mpixel is pretrained on text written in Latin (English), Cyrillic (Ukrainian), Simplified Chinese characters (Chinese), and Devanagari (Hindi). 
For each script, a corresponding subset of the mC4 \citep{xue-etal-2021-mt5} corpus is
rendered into images, following the strategy of rendering two characters per image patch from \citet{lotz-etal-2023-text}.
With a sequence length of 529 image patches and a batch size of 256, the model observes approximately 135 billion image patches over 1 million pretraining steps -- this is the same total amount of data as the original \pixel and \pixelbigrams models.
However, \mpixel is trained on an order-of-magnitude more unique samples than \pixelbigrams.
This difference is due to the fact that \pixelbigrams was trained by iterating 10 times over the English-only Wikipedia + BookCorpus datasets~\citep{Zhu_2015_ICCV}, whereas \mpixel processes each sample in our subset of mC4 only once across the four pretraining languages. processes each sample only once while still adhering to the same number of update steps.

\subsection{Pretraining Procedure}
\label{sec:method:pretraining}
Both \mpixel and \pixelbigrams follow the \pixel pretraining recipe from \citet{rust2023pixel}, including hyperparameter values.
Based on the Masked Autoencoding Vision Transformer \citep{he-etal-2022-masked}, the models render each input sequence to a 529-patch image using the PangoCairo rendering library,\footnote{\url{https://docs.gtk.org/PangoCairo}} where each image patch is $16\times 16$ pixels.
We use the Google Noto Sans fonts collection to ensure that the 
majority of Unicode codepoints can be accurately rendered.\footnote{\url{https://fonts.google.com/noto}}
\mpixel is trained by mixing the four languages within each batch; however, each individual sample consists of only one language.
The image patches are first embedded through a linear projection, 25\% of them are masked (in spans of up to 6 consecutive patches), and only the unmasked patches plus a \texttt{CLS} token are passed to the encoder.
After the encoder, a lightweight decoder reconstructs the pixel values of only the masked patches.
For downstream tasks we remove the decoder and instead attach a task-specific head, and disable patch masking in inputs.

\section{Experimental Setup}
\label{sec:experiments}

\begin{table*}

\renewcommand{\arraystretch}{1.12}
\centering

\begin{subtable}{\textwidth}
\centering
\resizebox{\textwidth}{!}{%
\begin{tabular}{lccccccccccc}
\toprule
& \multicolumn{3}{c}{Arabic}
& \multicolumn{5}{c}{Brahmic}
& \multicolumn{3}{c}{Cyrillic}\\
\cmidrule(lr){2-4}\cmidrule(lr){5-9}\cmidrule(lr){10-12}
& \arz & \uig & \urd
& \ben & \bod & \hin & \tam & \tel
& \kir & \rus & \ukr\\ \midrule
\monobert & 29.1 & 43.9 & 31.1
          & 38.4 & 40.7 & \textbf{87.2} & 48.6 & 29.5
          & \textbf{73.5} & \textbf{83.8} & \textbf{86.5}\\
\canine & \textbf{65.1} & 53.5 & \textbf{55.5}
          & \textbf{54.5} & 39.3 & 53.4 & 57.5 & 44.5 
          & 65.5 & 75.9 & 74.7 \\
\pixelbgs & 38.3 & 48.6 & 36.5
          & 31.7 & 36.9 & 32.6 & 39.7 & 39.9
          & 47.1 & 37.7 & 44.4\\
\mpixel   & 37.5 & \textbf{53.7} & 41.6
          & 46.2 & \textbf{46.3} & 78.6 & \textbf{64.5} & \textbf{46.6}
          & 62.9 & 74.7 & 80.5\\
\end{tabular}}%
\end{subtable}

\vspace{0.6ex}
\noindent\makebox[\textwidth]{\rule{\textwidth}{0.5pt}}

\begin{subtable}{\textwidth}
\centering
\resizebox{\textwidth}{!}{%
\begin{tabular}{lccccccccccccc}
& \multicolumn{6}{c}{Latin}
& \multicolumn{3}{c}{CJK}
& \multicolumn{3}{c}{Others}
& \multirow{2}{*}{Avg.}
\cr
\cmidrule(lr){2-7}
\cmidrule(lr){8-10}
\cmidrule(lr){11-13}
& \deu & \eng & \fin & \fra & \tur & \uzn
& \zho & \jpn & \kor & \greek & \heb & \hye & \\ \midrule
\monobert & 63.8 & \textbf{88.1} & 43.5 & 76.1 & 62.7 & 59.4
          & \textbf{89.5} & \textbf{78.9} & 15.4 & 32.6 & 32.7 & 36.5 & 55.3 \\
\canine & \textbf{77.7} & 78.4 & \textbf{61.6} & \textbf{77.4} & \textbf{68.2} & \textbf{63.6}
          & 77.9 & 75.3 & 64.5 & \textbf{65.3} & \textbf{54.4} & \textbf{60.8} & \textbf{63.7}\\
\pixelbgs & 63.8 & 84.3 & 59.7 & 73.2 & 60.7 & 56.7
          & 48.5 & 41.0 & 37.8 & 34.3 & 26.7 & 37.3 & 46.0 \\
\mpixel   & 67.3 & 83.9 & 60.6 & 70.7 & 59.9 & 56.2
          & 75.5 & 65.0 & \textbf{64.7} & 36.9 & 31.3 & 44.8 & 58.7 \\
\bottomrule
\end{tabular}}
\end{subtable}
\caption{%
Text classification results on a selected language subset of the SIB-200 benchmark using macro F1-score.
\monobert indicates that the monolingual \bert model varies by language (see \S\ref{sec:experiments:models}
for details).
Best performances are bolded.
\mpixel significantly outperforms its English-only-pretrained equivalent \pixelbigrams in almost all non-Latin languages, and \mpixel performs better than monolingual \bert models on novel writing systems.
Even though \canine is pretrained on 104 languages, \mpixel outperforms it on 8 of 23 languages.
}
\label{tab:results:sib}

\end{table*}

\subsection{Tasks \& Benchmarks}
\label{sec:experiments:tasks}

\paragraph{Text Classification.}
\label{sec:experiments:tasks:tc}
We first test the models on the sentence-level semantic task of topic classification using the SIB-200 benchmark~\citep{adelani-etal-2024-sib}.
Each example in SIB-200 is semantically aligned across languages.
This aspect of SIB-200 allows us to make a controlled comparison across different languages and scripts.
Our first set of evaluations cover the four pretraining languages of \mpixel: Latin (English \eng), Han (Chinese \zho), Cyrillic (Ukrainian \ukr), and Devanagari (Hindi \hin).
For the same-script transfer setting, we experiment with Latin script languages (German \deu, Finnish \fin, French \fra, Turkish \tur, Uzbek \uzn) and Cyrillic script languages (Kyrgyz \kir, Russian \rus).
For the related-script transfer setting, we perform experiments in Japanese (\jpn) and Brahmic script languages (Bengali \ben, Standard Tibetan \bod, Tamil \tam, Telugu \tel).
Lastly, we cover Armenian (\hye), Greek (\greek), Hebrew (\heb), Korean (\kor) and Arabic script languages (Egyptian Arabic \arz, Uyghur \uig, Urdu \urd) to test transfer to unrelated novel scripts.
We report macro-averaged F1 score as the metric.

\paragraph{Dependency Parsing.}
\label{sec:experiments:tasks:dp}
We evaluate on the token‐level syntactic parsing task of dependency parsing using the Universal Dependencies (UD) benchmark~\citep{nivre-etal-2020-universal,udp2.10}.
We also compare the models using the same three transfer learning settings again:
(i) same-script languages seen during pretraining: Latin (English \eng, Vietnamese \vie), Devanagari (Hindi \hin), Han (Chinese \zho), and Cyrillic (Ukrainian \ukr, Russian \rus, Bulgarian \bul);
(ii) languages in scripts related to at least one pretraining script: Coptic (\cop), Japanese (\jpn) and Brahmic script languages (Tamil \tam, Telugu \tel);
(iii) languages in scripts unrelated to the pretraining scripts: Arabic abjad (Arabic \ara, Urdu \urd) and Korean (\kor).
We report Labeled Attachment Score (LAS) as the evaluation metric.

\paragraph{Named Entity Recognition.}
\label{sec:experiments:tasks:ner}
Lastly, we perform experiments on the token-level semantic task of Named Entity Recognition (NER) using three benchmarks:
the multilingual Universal NER~\citep[UNER]{mayhew-etal-2024-universal} and Naamapadam~\citep{mhaske-etal-2023-naamapadam} benchmarks, as well as the NER portion of the Korean Language Understanding Evaluation~\citep[KLUE]{park_klue_korean_benchark}.
Once again, we cover same-script, related-script and unrelated-script transfer scenarios.
Here, three of the four scripts seen during pretraining -- Latin (English \eng, Serbian \srp), Han (Chinese \zho), and Devanagari (Hindi \hin) -- are additionally evaluated on Korean \kor, as well as three Brahmic scripts (Bengali \ben, Tamil \tam, Telugu \tel).
We report macro-averaged F1 scores.

\subsection{Baselines}
\label{sec:experiments:models}
We mainly compare \mpixel against the monolingual \pixelbigrams model, which is trained exclusively on English text rendered at the bigram level.
\mpixel implements the identical architecture, text rendering strategy and pretraining procedure with the same set of hyperparameters,
but \mpixel is multilingually pretrained on equal amounts of English, Hindi, Ukrainian and Simplified Chinese.
This comparison allows us to observe the effects of multilingual pretraining for pixel language models in different transfer learning settings.

We also compare \mpixel against four monolingual \bert variants:
The original English \bert~\citep{devlin-etal-2019-bert}
primarily for the Latin languages,
Chinese \bert~\citep{devlin-etal-2019-bert}
for Han and Japanese scripts,
a Hindi \bert~\citep{samuel-etal-2023-trained}
for the Brahmic script languages,
and a Ukrainian \bert~\citep{samuel-etal-2023-trained}
for the Cyrillic languages.
English \bert is also used as a fallback option to evaluate languages that do not belong to any of the pretraining scripts, such as Arabic or Hangul.
This allows us to test whether multilingually-pretrained pixel models can match or exceed the cross-lingual transfer capabilities of the tokenizer-based models, not only for Latin scripts but also for others.

Lastly, we include \canine \citep{clark-etal-2022-canine} as a character-level multilingual baseline, which is comparable in size to \mpixel.
However, \canine is pretrained with a subword loss objective on 104 languages using the same multilingual Wikipedia corpus as mBERT \citep{devlin-etal-2019-bert} whereas \mpixel has only seen four languages in pretraining.
Consequently, this difference in pretraining data and languages prevents a fully fair comparison of these models.

\begin{table*}[ht!]
\centering
\renewcommand{\arraystretch}{1.3}
\resizebox{\textwidth}{!}{%
\begin{tabular}{lllllllllllllllc}
\toprule
 &
  \multicolumn{2}{c}{Arabic} &
  \multicolumn{3}{c}{Brahmic} &
  \multicolumn{3}{c}{Cyrillic} &
  \multicolumn{2}{c}{Latin} &
  \multicolumn{3}{c}{CJK} &
  {\small Other} &
  \multirow{2}{*}{Avg.} \\ \cmidrule(lr){2-3} \cmidrule(lr){4-6} \cmidrule(lr){7-9} \cmidrule(lr){10-11} \cmidrule(lr){12-14} \cmidrule(lr){15-15}
          & \ara & \urd & \hin & \tam & \tel & \bul & \rus & \ukr & \eng & \vie & \zho & \jpn & \kor & \cop &      \\ \midrule
\monobert & \textbf{77.7} & 71.9 & \textbf{92.8} & 43.4 & 75.6 & \textbf{89.8} & \textbf{87.5} & \textbf{92.0} & \textbf{90.6} & \textbf{49.4} & \textbf{85.5} & 87.9 & 30.2 & 13.0 & 70.5 \\
\canine & 75.5 & 73.0 & 86.2 & 48.2 & 77.7 & 82.1 & 80.3 & 74.9 & 82.4 & 41.8 & 75.1 & \textbf{91.2} & 77.7 & 68.8 & 73.9 \\
\pixelbgs & \textbf{77.7} & 75.3 & 88.6 & 49.8 & 79.0 & 86.3 & 79.1 & 74.4 & 89.6 & \textbf{49.4} & 73.9 & 90.8 & 78.1 & 81.4 & 76.7 \\
\mpixel   & 74.2 & \textbf{75.9} & 91.6 & \textbf{60.5} & \textbf{79.7} & 88.8 & 83.0 & 84.3 & 87.6 & \textbf{49.4} & 79.9 & \textbf{91.2} & \textbf{82.3} & \textbf{81.6} & \textbf{79.3} \\ \bottomrule
\end{tabular}%
}
\caption{%
Dependency parsing results for the selected set of languages in the \textsc{udp} benchmark with LAS.
\monobert indicates that the monolingual \bert model varies by language.
\mpixel outperforms \pixelbigrams in non-Latin script languages, and it again achieves a better performance than \monobert on novel scripts, while showing substantial gains over \canine on every language but one.
}
\label{tab:results:udp}
\end{table*}

\begin{table}[ht]
\renewcommand{\arraystretch}{1.3}
\centering
\resizebox{\columnwidth}{!}{%
\begin{tabular}{llllllllll}
\toprule
 & \multicolumn{2}{c}{Latin} & \multicolumn{4}{c}{Brahmic} & \multicolumn{2}{c}{CJK}                          & \multirow{2}{*}{Avg.} \\ \cmidrule(lr){2-3} \cmidrule(lr){4-7} \cmidrule(lr){8-9}
                       & \eng        & \srp        & \hin  & \ben  & \tam & \tel & \multicolumn{1}{c}{\kor} & \multicolumn{1}{c}{\zho} &                       \\ \midrule
\monobert    & \textbf{79.3} & \textbf{85.8} & \textbf{82.5} & 75.4 & 67.3 & 78.3 & 30.6 & \textbf{85.4} & 73.1 \\
\canine    & 67.1 & 83.0 & 79.5 & \textbf{78.5} & \textbf{69.7} & \textbf{80.5}  & 80.6 & 75.8 & \textbf{76.8} \\
\pixelbgs & 63.4 & 81.6 & 79.0 & 78.0 & 67.9 & 79.6 & 80.4 & 61.4 & 73.9 \\
\mpixel   & 67.3 & 82.1 & 80.9 & \textbf{78.5} & 68.0 & 79.6 & \textbf{81.6} & 74.9 & 76.6 \\ \bottomrule
\end{tabular}%
}
\caption{%
NER results by macro-averaged F1-scores.
\monobert is the monolingual \bert model, which varies by language depending on script. Overall, \mpixel performs on par with \canine, and outperforms \pixelbgs and \monobert with an average score of $76.6$.}

\label{tab:results:ner}
\end{table}

\section{Results and Discussion}
\label{sec:experiments:results}

\paragraph{Text Classification.}
Table \ref{tab:results:sib} presents the results on SIB-200 for text classification.
\mpixel outperforms \pixelbigrams by large margins in its pretraining languages (\hin: +46.0, \ukr: +36.1, \zho: +27.0), which are unseen by \pixelbigrams during the pretraining.
We also observe substantial gains in Cyrillic languages (\kir: +15.8, \rus: +37.0),
showing that pretraining pixel models on a particular script enhances transfer learning within the same-script languages.
In English and other Latin languages, both models achieve similar performances.
The significant performance gains in Japanese (\jpn: +24) and the Brahmic languages (\ben: +14.5, \bod: +9.4, \tam: +24.8, \tel: 6.7) showcase \mpixel's cross-lingual transfer learning ability to novel scripts orthographically related to one pretraining script.
Lastly, we compare both \mpixel and \pixelbigrams in languages with writing systems visually distant to the pretraining scripts.
Once again, \mpixel performs
better than \pixelbigrams in these languages,
where we can observe improvements
for Armenian (\hye: +7.5), Greek (\greek: +4.3), Korean (\kor: +26.9) and the languages in right-to-left abjad writing systems (\heb: 4.6, \uig: +5.1, \urd: +5.1).
These results
illustrate that multilingual pretraining with a diverse set of scripts accelerates cross-lingual generalization even for novel and distant writing systems.
Overall, these results highlight that visually and linguistically diverse multilingual pretraining for pixel models leads to substantial gains in all types of transfer learning scenarios investigated in this work.

Compared to the monolingual \bert variants, \mpixel performs consistently better, especially in the transfer learning setting involving unseen scripts.
Conversely, \monobert models surpass \mpixel in transfer learning within the same-script,
yet \monobert pretrained in English falls behind \mpixel in German (\deu: +3.5) and Finnish (\fin: +17.1).

We further compare \mpixel against character-level multilingually-pretrained \canine.
\mpixel outperforms \canine on 3 out of 4 pretraining languages and 4 out of 5 Brahmic languages, as well as slight gains in \kor and \uig.
These gains are notable given that \canine was pretrained on all of these languages except \arz and \bod.

\paragraph{Dependency Parsing.}
Table \ref{tab:results:udp} presents the results on the UDP benchmark. %
In the pretraining languages, \mpixel significantly improves upon \pixel (\hin: +3.0, \ukr: +9.9, \zho: +6.0)
except in English (\eng: -2.0), which both models have seen in their pretraining.
\mpixel outperforms \pixel on the languages written in Cyrillic (\bul: +2.5, \rus: +3.9),
which demonstrates improved cross-lingual transfer learning within the same-script languages once again.
For the unseen Brahmic languages, \mpixel achieves a slight gain in Telugu (\tel: +0.7) and a much larger performance boost in Tamil (\tam: +10.7).
For the orthographically distant Korean language, \mpixel outperforms \pixelbigrams (\kor: +4.2).
For the Arabic‐script languages, we observe mixed results: In Arabic, the performance drops (\ara: -3.5), while we observe a modest gain in Urdu (\urd: +0.6).
Altogether, multilingually-pretrained \mpixel improves on \pixelbigrams on the dependency parsing task for the unseen languages considering various cross-lingual transfer learning settings.
Our findings on this task is similar to the SIB-200 findings for comparing \mpixel against monolingual \bert models:
(i) \mpixel achieves a better overall performance than \monobert in cross-lingual transfer involving writing systems unknown to both;
(ii) \monobert performs better than \mpixel for the pretraining scripts
and cross-lingual transfer within the same-script. %
Lastly, when compared with \canine, \mpixel achieves consistently better overall performance, despite the fact that \canine has been trained on all languages except \cop.

\begin{figure*}[!ht]
  \centering
  \includegraphics[width=\textwidth]{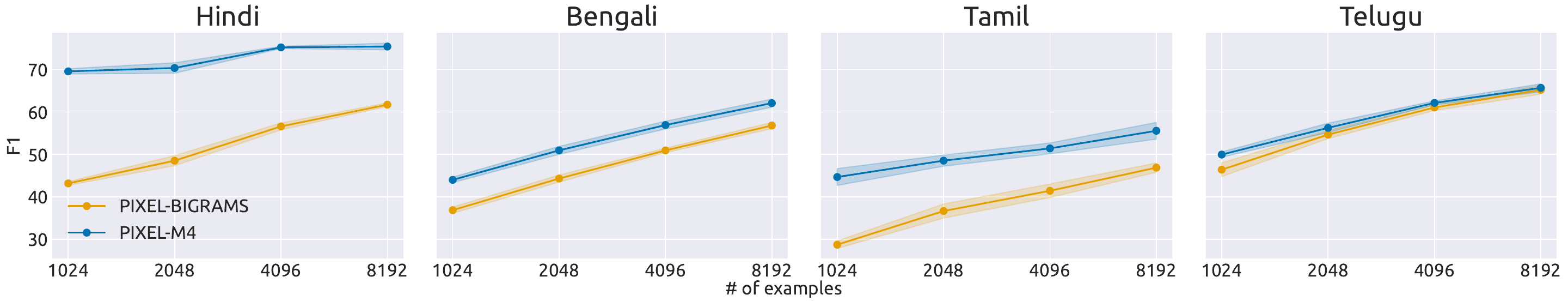}
  \caption{
    Data‐efficient learning experiments on the Naamapadam NER benchmark showing the mean test F$_1$ score as a function of training set size in log scale for four Brahmic languages. In each experiment, \mpixel consistently outperforms \pixelbigrams, with the largest relative gains under the smallest data regimes.}
  \label{fig:ner-data-efficiency}
\end{figure*}

\paragraph{Named Entity Recognition.}
Table \ref{tab:results:ner} reports macro-averaged F1 for NER across eight languages.
As expected, multilingual pixel pretraining (\mpixel) outperforms the English‐only \pixelbigrams model on every language, raising the average F1 from 73.9 to 75.9.
The largest boost is seen in Chinese (\zho: +13.5), reflecting that exposure to Chinese during \textsc{pixel-m4}’s pretraining.
Other pretraining languages also benefit from multilingual pretraining (\eng: +3.9, \hin: +1.9).
Differently from the other tasks, both \mpixel and \pixelbigrams perform on par in the Brahmic scripts (\hin: +1.9, \ben: +0.5, \tam: +0.1, \tel: 0.0):
This might be due the larger training sets available in the Naamapadam benchmark. 
Later, in \S\ref{sec:analysis}, we show that \mpixel outperforms \pixelbigrams with large margins in low-resource settings.
Lastly, +1.2 gain in Korean suggests that \mpixel can transfer visual substructure from unrelated scripts for better entity processing.

The monolingual \bert models achieve a better performance than \mpixel for the languages with writing systems known by both models, underscoring that world‐knowledge and semantic co‐occurrence patterns encoded into specific token entities remain crucial for this semantic task.
This is especially the case for English, as both \bert and \pixelbigrams are pretrained using exactly the same data.
Nonetheless, our findings for the languages in unseen scripts is inline with previous experiments
where \mpixel performs better than \monobert: (\ben: +3.1, \tam: +0.7, \tel: 1.6).
These improvements highlight how pixel models can process languages in related scripts directly, avoiding the tokenization failure modes of subword-based models.

Finally, \mpixel performs on par with \canine overall, although \canine was exposed to all downstream task languages during pretraining.

\section{Analysis}
\label{sec:analysis}

\subsection{Data-Efficiency Analysis}
To investigate the capabilities of \mpixel further, we perform a data-efficiency analysis on Naamapadam -- the Indic languages benchmark. 
Using the original training splits, we create subsets of size 1024, 2048, 4096 and 8192 examples.
We repeat this process 8 times using different random seeds, resulting 32 different subsets.
Next, we train both \pixelbigrams and \mpixel on these subsets and compare them
in terms of data-efficiency.
Figure \ref{fig:ner-data-efficiency} illustrates this comparison, where each subplot represents the results for the specified language.
For Hindi, Bengali and Tamil, \mpixel performs significantly better than \pixel in all settings.
The results in Bengali and Tamil also highlight the cross-lingual transfer learning capacity of the \mpixel in low-resource settings.
As we decrease the number of examples, we observe more substantial gains in all languages including Telugu, where \mpixel performs slightly better than \pixelbigrams on the entire set of tasks.
Overall, multilingual pretraining of pixel language models substantially enhances transfer learning in low-resource settings.

\begin{figure*}[!ht]
  \centering
  \includegraphics[width=\textwidth]{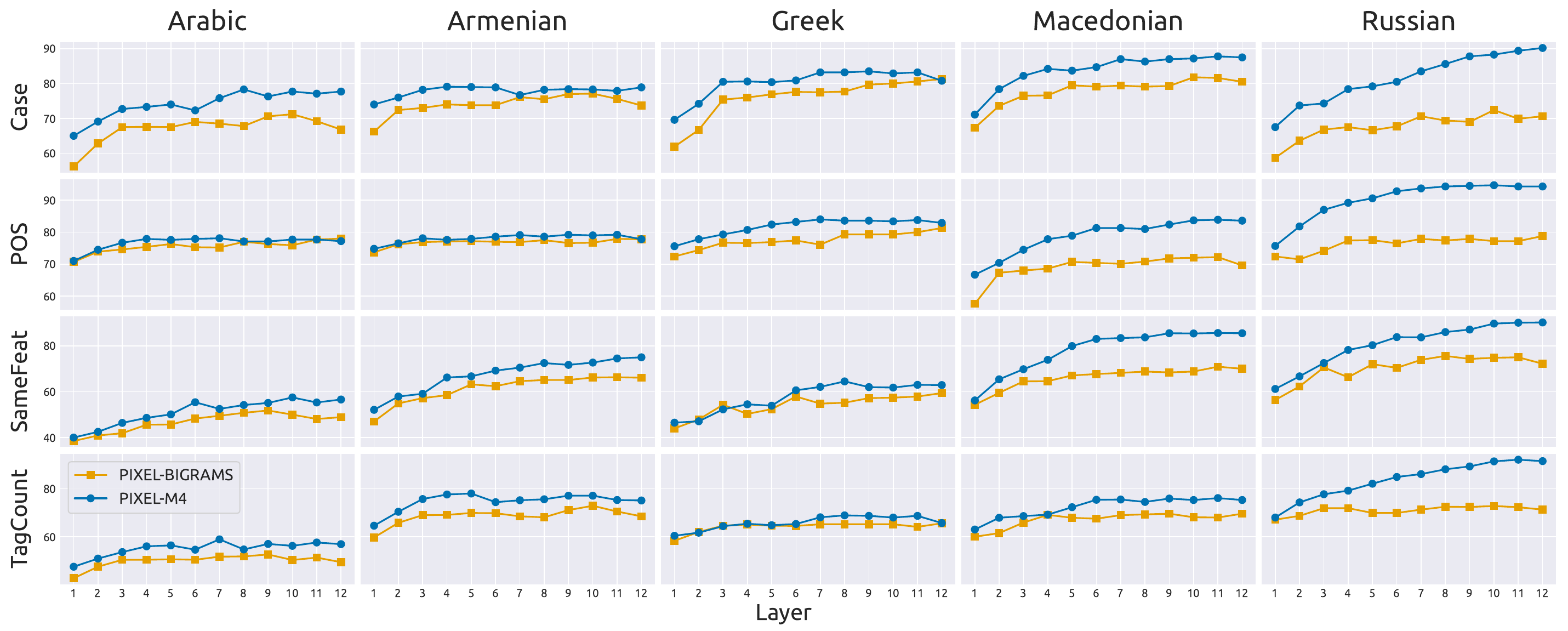}
  \caption{
    Word-level probing analysis on \textsc{linspector}, where each row investigates a different task, and each column investigates a different language. In each subplot, y-axis represents the model accuracies and x-axis represents the corresponding layer number for the used hidden representations.
    Multilingually-pretrained \mpixel has learned better linguistic representations even for the languages with orthographically distant writing systems.}
  \label{fig:linspector-main}
\end{figure*}

\begin{figure*}[!ht]
  \centering
  \includegraphics[width=\textwidth]{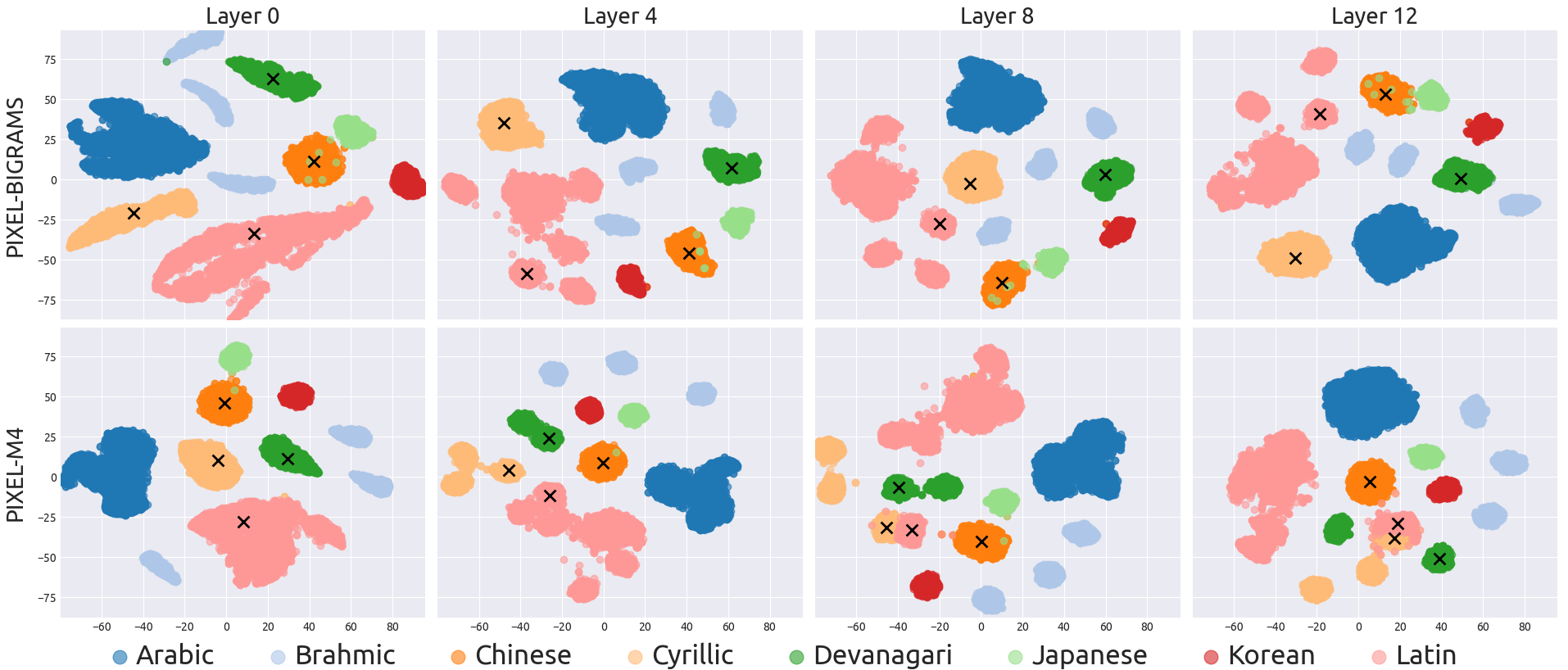}
  \caption{
  t-SNE visualization of the outputs for the specified layers. Each row contains visualizations for a particular model, and each column focuses on a particular layer. Each `$\bm{\times}$' marker appear at the centroid of a different pretraining language seen by \mpixel. Both models cluster languages based on their scripts, yet \mpixel clusters some pretraining languages in the later layers.
  }
  \label{fig:tsne-main}
\end{figure*}

\subsection{Word-Level Probing}
We also performed a probing analysis similar to \citet{tatariya-etal-2024-pixology}.
Here, we use \textsc{linspector} \citep{csahin-etal-2020-linspector}, a multilingual word-level probing benchmark, to investigate the transferability of multilingual representations encoded by \mpixel.
We investigate hidden representations encoded by both \mpixel and \pixelbigrams after each layer, and compare them against each other.
We perform this analysis on four different tasks (Case Marking, POS, SameFeat, TagCount) using five different languages (Arabic, Armenian, Greek, Russian, Macedonian).\footnote{See Appendix for a larger set of tasks and languages.}
\textit{Case Marking} requires assessing the grammatical case (e.g. nominative, accusative) of a given input word.
\textit{POS} involves predicting the POS tag for the given word.
The \textit{SameFeat} task measures the ability to detect the mutual morphological feature of two given words in their surface forms.
Lastly, \textit{TagCount} requires correctly predicting the number of morphological tags for the given input word.
\textit{SameFeat} and \textit{TagCount} are more difficult than the other tasks, as both require predicting the entire set of morphological features for the given word(s).

We show the results of our probing analyses in Figure \ref{fig:linspector-main}.
In this grid of subplots, each row investigates a different task, and each column investigates a different language.
In Macedonian and Russian, \mpixel learns significantly better representations compared to \pixelbigrams, which is expected because \mpixel has seen a similar language in the same script during pretraining.
The gap between two models in earlier layers (1-3) is smaller on \textit{SameFeat} and \textit{TagCount}, as they require more complex linguistic assessment.
This also applies for the other tested languages, and it is in line with the observations of \citet{tatariya-etal-2024-pixology}, where earlier layers focus more on visual rather than semantic processing.
In Arabic, Armenian, and Greek, \mpixel still performs slightly better than \pixelbigrams on the majority of tasks, which showcases its improved visual processing and transfer learning to unseen languages.
For these unseen languages, the performance of \mpixel starts to plateau starting from the 7th or 8th layer.
Overall, these results demonstrate that the multilingual pretraining produces a better set of hidden representations throughout the entire model, even for the unseen scripts.

\begin{figure}[!ht]
  \centering
  \includegraphics[width=\columnwidth]{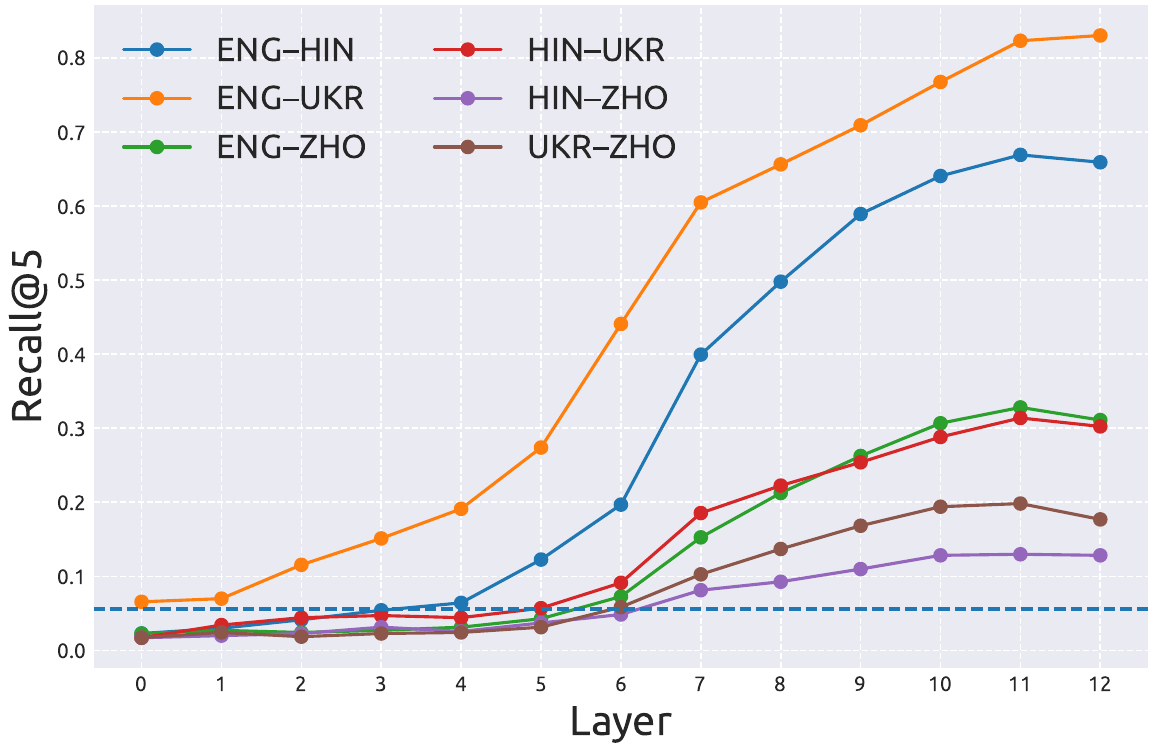}
  \caption{
  Cross-lingual similarity analysis on SIB-200 using the mean pooled hidden representations of \mpixel. The x-axis indicates the layer number; the y-axis reports the performance using recall@5. Each line focuses on a different language-pair combination.
  The dashed line shows the maximum recall@5 value obtained by \pixelbigrams for these language pairs.
  This analysis reveals that \mpixel has learned a mutual semantic representation for some pretraining language pairs. 
  }
  \label{fig:sib200-retrieval}
\end{figure}

\subsection{Analyzing Hidden Representations}
Similar to \citet{salesky-etal-2023-multilingual}, we visualize the hidden representations learned by both \pixelbigrams and \mpixel using t-SNE \citep{van2008visualizing}.
To perform this analysis, we use a subset of SIB-200 \citep{adelani-etal-2024-sib} including the training splits of 26 languages.
We perform t-SNE visualization throughout the model, starting from the convolved input representations (Layer 0) to the output of the last transformer layer (Layer 12).
Figure 4 shows t-SNE plots: rows correspond to models, columns to layers, and ‘×’ marks the \mpixel pretraining-language centroids.
We observe the same phenomenon for the convolved features as demonstrated in \citet{salesky-etal-2023-multilingual}:
Languages which use the same or a related writing script are grouped together.
This can be observed for both models, where we can see large clusters for Arabic, Cyrillic and Latin, and Chinese-Japanese language clusters appear next to each other.
As we move through in the model layers, we start to see some languages form their own separate clusters by moving away from their script clusters (e.g. Layer 4 and 8).
More importantly, in the later layers of \mpixel, we observe that the pretraining languages move away from the rest of the languages that share the same script, and they start to cluster together.
This observation demonstrates that \mpixel shifts its focus from visual processing more to the semantics in the later layers.
This raises the question of whether \mpixel has learned a semantic representation space shared between different pretraining languages.

To determine whether \mpixel has learned a representation space shared between different pretraining languages, we perform a cross-lingual retrieval experiment on the multilingually aligned SIB-200 benchmark.
To obtain sentence embeddings, we apply L2 normalization to the mean pooled hidden representations after each layer.
At each layer, we treat each sentence embedding in one language as a query and compute its cosine similarity against every sentence embedding in the other language.
We report recall@5, i.e., the percentage of the examples where the true translation is ranked in the top 5.
Since each sentence has exactly one correct translation, retrieval performance per example is binary, taking values of either 0 or 1.
Figure \ref{fig:sib200-retrieval} shows the results for each language pair.
We see that the semantic alignment between each language pair increases as we move through in the layers.
Particularly, the semantic alignment between English and Ukrainian is very high, as they are also tightly clustered in the t-SNE feature space.
We can also observe a high semantic alignment between English and Hindi, yet the remaining pairs do not share a highly aligned semantic representation space.

\section{Related Work}
\label{sec:related_work}
\citet{salesky-etal-2021-robust} proposed an encoder-decoder-based machine translation model that replaces the tokenizer in the encoder by processing source text as rendered images.
\citet{rust2023pixel} proposed \pixel, the first model that relies on purely processing visually rendered text.
Later, \citet{lotz-etal-2023-text} investigated different strategies for text rendering with the aim of removing redundant patches.
\citet{fei-etal-2024-mtls} experimented with replacing \bert's tokenizer with pixel-based processing.
\citet{gao-etal-2024-screenshots} extended \pixel with a mixed modality pretraining objective, which produced substantial improvements.
\citet{tai-etal-2024-pixar} pretrained \textsc{pixar}, which is the first autoregressive pixel language model that purely relies on processing rendered text.
\citet{gao-etal-2024-screenshots,chai-etal-2024-autoregressive} also proposed pixel language models with text generation abilities, yet they achieved this by still depending on subword tokenizers.
Recently, \citet{lotz-etal-2025-overcoming} embedded pixel language models into the English-centric language models as a fallback mechanism to better adapt these models to novel languages and scripts.
Most notably, \citet{salesky-etal-2023-multilingual} is closely related to our work as it employs a multilingual pretraining.
However, their experiments focus on learning a shared encoder for machine translation,
while we pretrained a multilingual pixel language model for general-representation learning without relying on any tokenizer.

\section{Conclusion}
\label{sec:conclusion}
In this work, we explored multilingual pretraining for pixel language models.
We pretrained \mpixel, a multilingual pixel-based language model on four visually and linguistically diverse languages, namely English, Hindi, Ukrainian and Simplified Chinese.
We performed downstream task experiments on three different tasks: sentence classification, dependency parsing, and named entity recognition.
In these experiments, we covered a diverse set of languages and scripts, where we evaluated on 27 languages and 15 scripts.
Our experiments revealed that \mpixel achieves superior performance in low-resource settings compared to its monolingually-pretrained predecessor \pixelbigrams,  outperforming it in almost all non-Latin languages by a large margin.
In order to better understand the representations learned by \mpixel, we conducted word-level and sentence-level analyses.
Our word-level probing analysis 
illustrated that \mpixel has learned better hidden representations than \pixelbigrams throughout the network for the unseen scripts, highlighting its cross-lingual transfer capabilities.
Additionally, an analysis on the hidden layer representations revealed that \mpixel has learned a semantic representation space shared by a subset of pretraining languages in the later layers.
In future work, we aim to scale up multilingual pretraining for pixel models with larger model capacity and more languages included in pretraining.

\section*{Limitations}
\label{sec:limitations}
\mpixel inherits many of the limitations of its predecessors.
First, rendering text using the bigrams strategy leads to increased sequence lengths when a bigram does not fit into single patch.
Like \citet{rust2023pixel} and \citet{lotz-etal-2023-text}, \mpixel cannot generate text.
The improvements over \pixelbigrams are also limited for Latin‐script languages and also for high-resource settings.
Due to our limited compute budget, we pretrained a single \mpixel model on only four languages-each in a different script.
Consequently, we have not explored larger or different combinations of languages and scripts, such as additional Latin-script languages (e.g. French, Estonian, Turkish) or right-to-left scripts (e.g. Hebrew, Arabic).
Finally, the comparison between \mpixel and \pixelbigrams is not entirely fair, as \mpixel was exposed to more data.
Nonetheless, pretraining another monolingual model on an equivalent amount of data for such a comparison was not feasible within our computational budget.
We leave these comparisons to future work.

\section*{Acknowledgements}
IK, IZ and DE were supported by the European Union’s Horizon 2020 research and innovation program under grant agreement No.\ 101135671 (TrustLLM). 
JFL is funded by the ROCKWOOL Foundation (grant 1242).
DE was supported by a research grant (VIL53122) from VILLUM FONDEN.
PR is funded by the Novo Nordisk Foundation (grant NNF 20SA0066568).

IK and IZ acknowledge the EuroHPC Joint Undertaking for awarding access to MareNostrum5, hosted at Barcelona Supercomputing Center (BSC), Spain, under proposals No. EHPC-DEV-2024D11-047 and EHPC-DEV-2024D12-031.

\bibliography{custom}

\appendix

\section{Appendix}
\label{sec:appendix}
This appendix section contains a summary of data statistics, implementation details of the downstream task experiments and the rest of the \textsc{LINSPECTOR} word-level probing analyses.

\subsection{Data Statistics}
We summarize data statistics of the benchmark used in this work in this section.
Table \ref{tab:appendix:sib-linspector-stats} contains statistics for SIB-200 \citep{adelani-etal-2024-sib,goyal-etal-2022-flores,nllb2022} and \textsc{linspector} \citep{csahin-etal-2020-linspector}, where each language split contains same number examples for training, validation and testing purposes.
Table \ref{tab:appendix:udp-statistics} reports the statistics of dependency parsing treebanks used in this work.
Lastly, we share the NER benchmarks statistics in Table \ref{tab:appendix:ner-statistics}.

\subsection{Implementation Details}
\noindent \textbf{\mpixel.} Table \ref{tab:hyperparams} lists the hyperparameter configurations used for pixel language models, \mpixel and \pixelbigrams, across downstream tasks.
Overall, we use the same set of hyperparameters with the previous work \citep{lotz-etal-2023-text}.
We repeat the same experiment using different random seeds.
For reporting test results, we average the test scores of the five runs with the highest validation split performance.

\noindent \textbf{\canine.}
We use the same experimental setup and hyperparameters for \canine with \mpixel and \pixelbigrams.

\noindent \textbf{Monolingual BERT Models.}
All models were fine-tuned in 16-bit BrainFloat~\citep{abadi2016tensorflowlargescalemachinelearning} using AdamW~\citep{kingma_adam, loshchilov_adamw} with a maximum learning rate of \num{5e-5} that is warmed up over the first 100 steps and subsequently linearly decayed toward 0.
Across all tasks, we fine-tune for at maximum 15,000 steps, while evaluating every 500 steps for dependency parsing and NER, whereas topic classification is evaluated every epoch.
Early stopping of 5 evaluation cycles (DP and NER) or 20 epochs with a threshold of 0.0 is implemented.
For all tasks and languages, when a separate evaluation split is available, we selected the checkpoint performing best on it and evaluated on the test split.
If no separate evaluation split was available, we selected and reported the best performance on the evaluation split.
Inputs were truncated or padded to a maximum length of 256 tokens for parsing and classification, and 196 tokens for NER.
For parsing and NER, a batch size of 64 is used, while topic classification is trained with batch size 32.
We followed~\citet{rust2023pixel} and evaluated dependency parsing using a biaffine parsing head~\citep{DBLP:conf/iclr/DozatM17,glavas-vulic-2021-supervised}.

\begin{table}[t]
\centering
\resizebox{\columnwidth}{!}{%
\begin{tabular}{@{}llrrr@{}}
\toprule
Benchmark & License & Train & Validation & Test \\ \midrule
SIB-200  & CC BY-SA 4.0 & 701   & 99         & 204  \\
LINSPECTOR & Apache 2.0 & 7000  & 2000       & 1000 \\ \bottomrule
\end{tabular}%
}
\caption{Data statistics for the equally-sized SIB-200 and \textsc{linspector} language splits.}
\label{tab:appendix:sib-linspector-stats}
\end{table}

\begin{table*}[ht]
\centering
\resizebox{0.6\textwidth}{!}{%
\begin{tabular}{@{}llrl@{}}
\toprule
Language & Treebank           & \#Sentences                  & License         \\ \midrule
\eng     & English-EWT        & 16621                        & CC BY-SA 4.0    \\
\ara     & Arabic-PADT        & 7664                         & CC BY-NC-SA 3.0 \\
\bul     & Bulgarian-BTB      & 11138 & CC BY-NC-SA 3.0 \\
\cop     & Coptic-Scriptorium & 2011                         & CC BY 4.0       \\
\hin     & Hindi-HDTB         & 16647                        & CC BY-NC-SA 4.0 \\
\jpn     & Japanese-GSD       & 8100                         & CC BY-SA 4.0    \\
\kor     & Korean-GSD         & 6339                         & CC BY-SA 4.0    \\
\rus     & Russian-GSD        & 5030                         & CC BY-SA 4.0    \\
\tam     & Tamil-TTB          & 600                          & CC BY-NC-SA 3.0 \\
\tel     & Telugu-MTG         & 5130                         & CC BY-SA 4.0    \\
\ukr     & Ukrainian-IU       & 5030                         & CC BY-NC-SA 4.0 \\
\urd     & Urdu-UDTB          & 5130                         & CC BY-NC-SA 4.0 \\
\vie     & Vietnamese-VTB     & 3000                         & CC BY-SA 4.0    \\
\zho     & Chinese-GSD        & 4997                         & CC BY-SA 4.0    \\ \bottomrule
\end{tabular}%
}
\caption{Total number of sentences of Universal Dependencies v2.10 \citep{udp2.10,nivre-etal-2020-universal} treebanks used for dependency parsing task evaluations, including dataset licenses. Adapted from \citet{rust2023pixel}.}
\label{tab:appendix:udp-statistics}
\end{table*}

\begin{table*}[ht]
\centering
\resizebox{0.6\textwidth}{!}{%
\begin{tabular}{@{}llrl@{}}
\toprule
Language & Source           & \#Sentences                  & License         \\ \midrule
\eng     & English-EWT        & 16621                        & CC BY-SA 4.0    \\
\srp     & Serbian-SET          & 4384 & CC BY-SA 4.0 \\
\hin     & Naamapadam & 1M                        & CC0 \\
\ben     & Naamapadam  & 967k & CC0 \\
\tam     & Naamapadam  & 501k & CC0 \\
\tel     & Naamapadam  & 511k & CC0 \\        
\kor     & KLUE        & 26k & CC BY-SA 4.0    \\
\zho     & Chinese-GSD & 4997 & CC BY-SA 4.0    \\ \bottomrule
\end{tabular}%
}
\caption{Overview of NER datasets \citep{mayhew-etal-2024-universal,mhaske-etal-2023-naamapadam,park_klue_korean_benchark}.}
\label{tab:appendix:ner-statistics}
\end{table*}

\begin{table*}[htbp]
\centering
\resizebox{0.6\textwidth}{!}{%
\begin{tabular}{@{}lccc@{}}
\toprule
Parameter            & SIB-200 & UDP     & NER     \\ \midrule
Classification head pooling & Mean       & ---     & ---     \\
Optimizer                   & \multicolumn{3}{c}{AdamW}      \\
Adam $\beta$                & \multicolumn{3}{c}{\num{0.9}, \num{0.999}} \\
Adam $\varepsilon$          & \multicolumn{3}{c}{\num{1e-8}}   \\
Weight decay                & \multicolumn{3}{c}{\num{0}}          \\
Learning rate          & \multicolumn{3}{c}{$\{\num{1e-5}, \num{3e-5}, \num{5e-5}, \num{7e-5}, \num{9e-5}\}$}\\

Learning rate schedule & \multicolumn{3}{c}{Linear decay}  \\
Warmup steps  & \multicolumn{3}{c}{\num{100}}        \\
Max sequence length         & \num{256}        & \num{256}     & \num{196}     \\
Stride                      & ---        & ---     & ---     \\
Batch size                  & \num{32}         & \num{64}      & \num{64}      \\
Max steps                   & \num{15000}      & \num{15000}   & \num{15000}   \\
Eval strategy               & epochs     & steps   & steps   \\
Eval steps                  & ---          & \num{500}     & \num{500}     \\
Early stopping              & \multicolumn{3}{c}{\checkmark} \\
Early stopping patience     & \num{20}         & \num{5}       & \num{5}       \\
Dropout probability         & \multicolumn{3}{c}{\num{0.1}}        \\ \bottomrule
\end{tabular}%
}
\caption{Hyperparameters used for fine-tuning and evaluating models on the SIB-200, UDP parsing, and NER tasks.}
\label{tab:hyperparams}
\end{table*}

\subsection{LINSPECTOR Results}
\begin{figure*}[htbp]
  \centering
  \includegraphics[width=\textwidth]{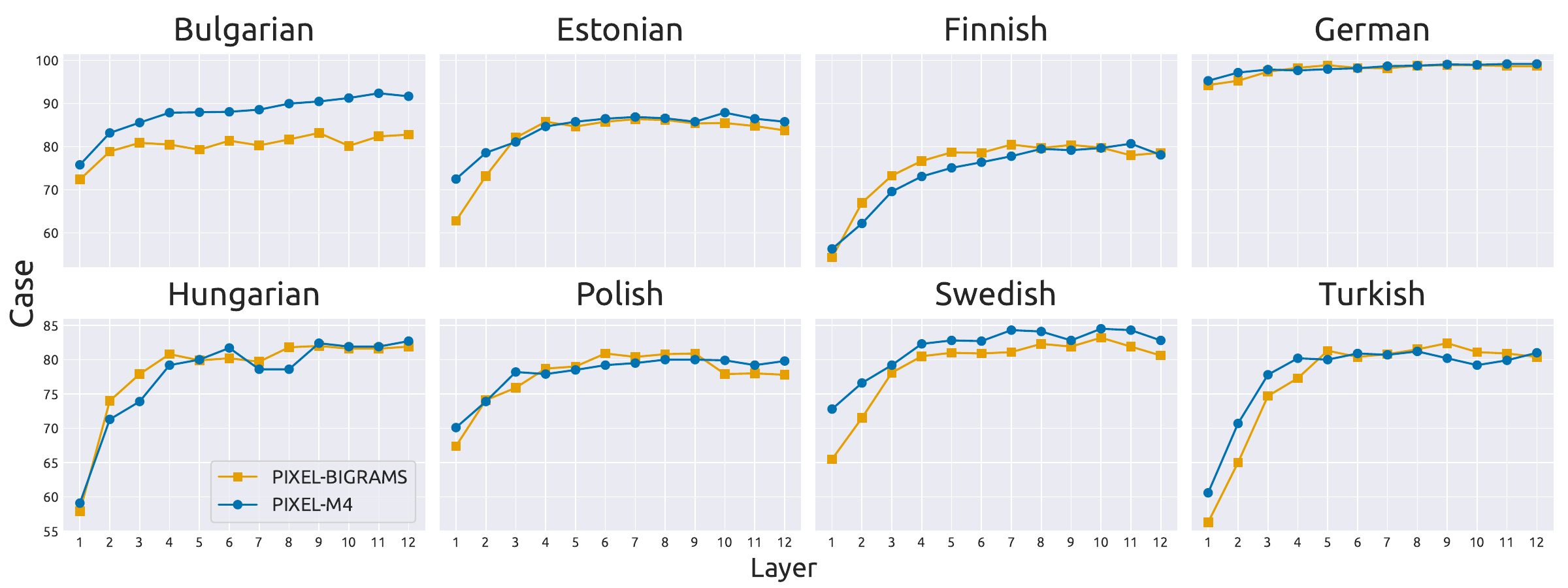}
  \caption{Word-level probing analysis on \textsc{linspector} for the Case task. Each subplot shows a different language; in each, the y-axis represents model accuracies and the x-axis represents layer number of the hidden representations.}
  \label{fig:linspector-appendix-case}
\end{figure*}

\begin{figure*}[htbp]
  \centering
  \includegraphics[width=\textwidth]{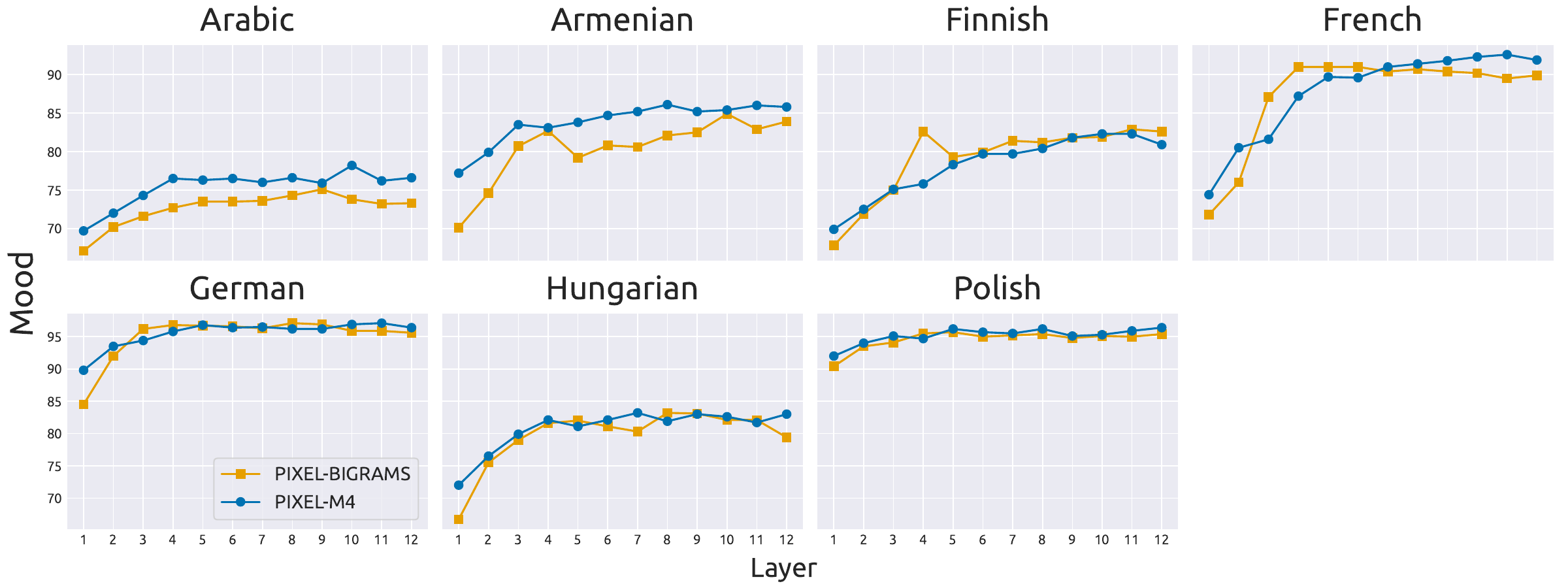}
  \caption{Word-level probing analysis on \textsc{linspector} for the Mood task. Each subplot shows a different language; in each, the y-axis represents model accuracies and the x-axis represents layer number of the hidden representations.}
  \label{fig:linspector-appendix-mood}
\end{figure*}

\begin{figure*}[htbp]
  \centering
  \includegraphics[width=\textwidth]{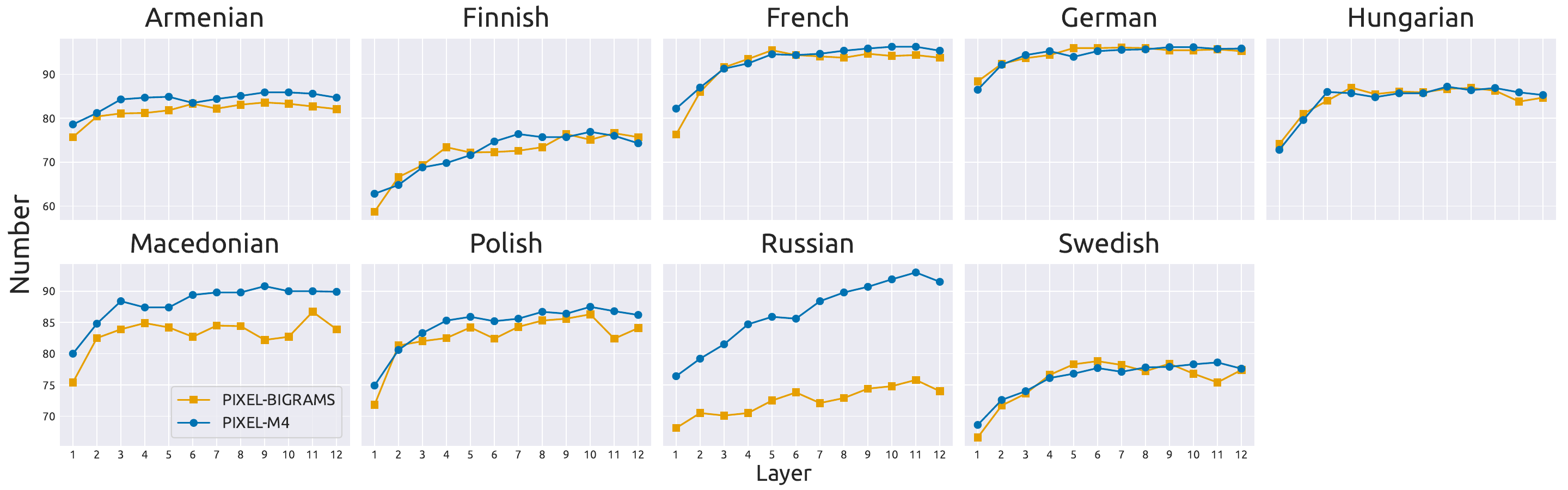}
  \caption{Word-level probing analysis on \textsc{linspector} for the Number task. Each subplot shows a different language; in each, the y-axis represents model accuracies and the x-axis represents layer number of the hidden representations.}
  \label{fig:linspector-appendix-number}
\end{figure*}

\begin{figure*}[htbp]
  \centering
  \includegraphics[width=\textwidth]{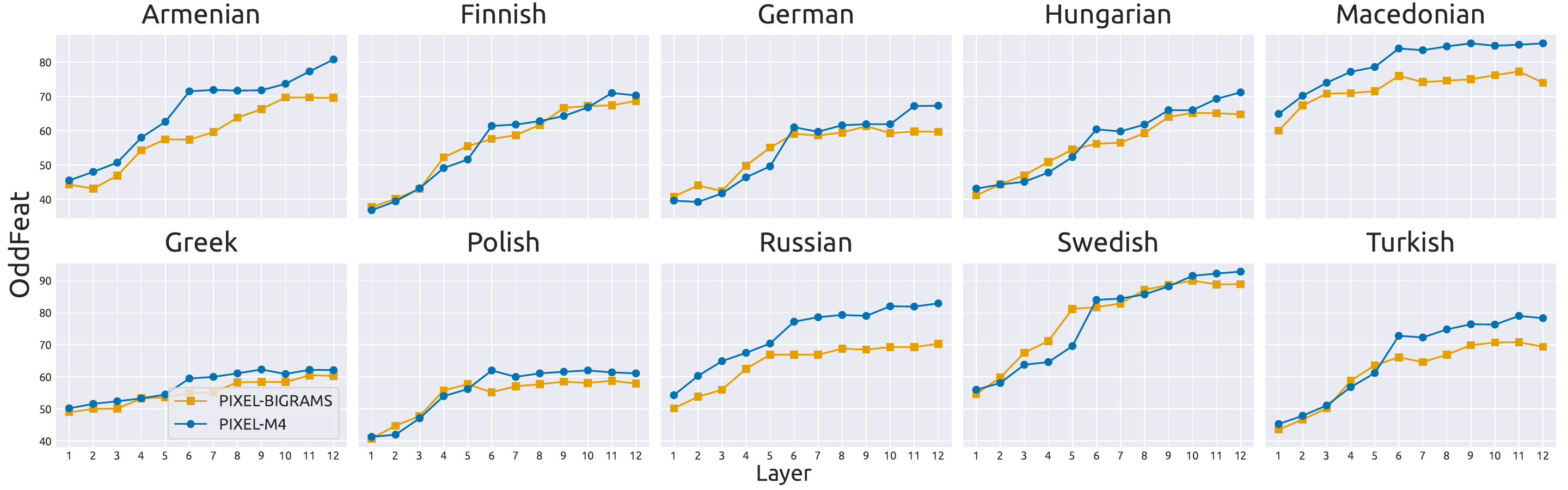}
  \caption{Word-level probing analysis on \textsc{linspector} for the OddFeat task. Each subplot shows a different language; in each, the y-axis represents model accuracies and the x-axis represents layer number of the hidden representations.}
  \label{fig:linspector-appendix-oddfeat}
\end{figure*}

\begin{figure*}[htbp]
  \centering
  \includegraphics[width=\textwidth]{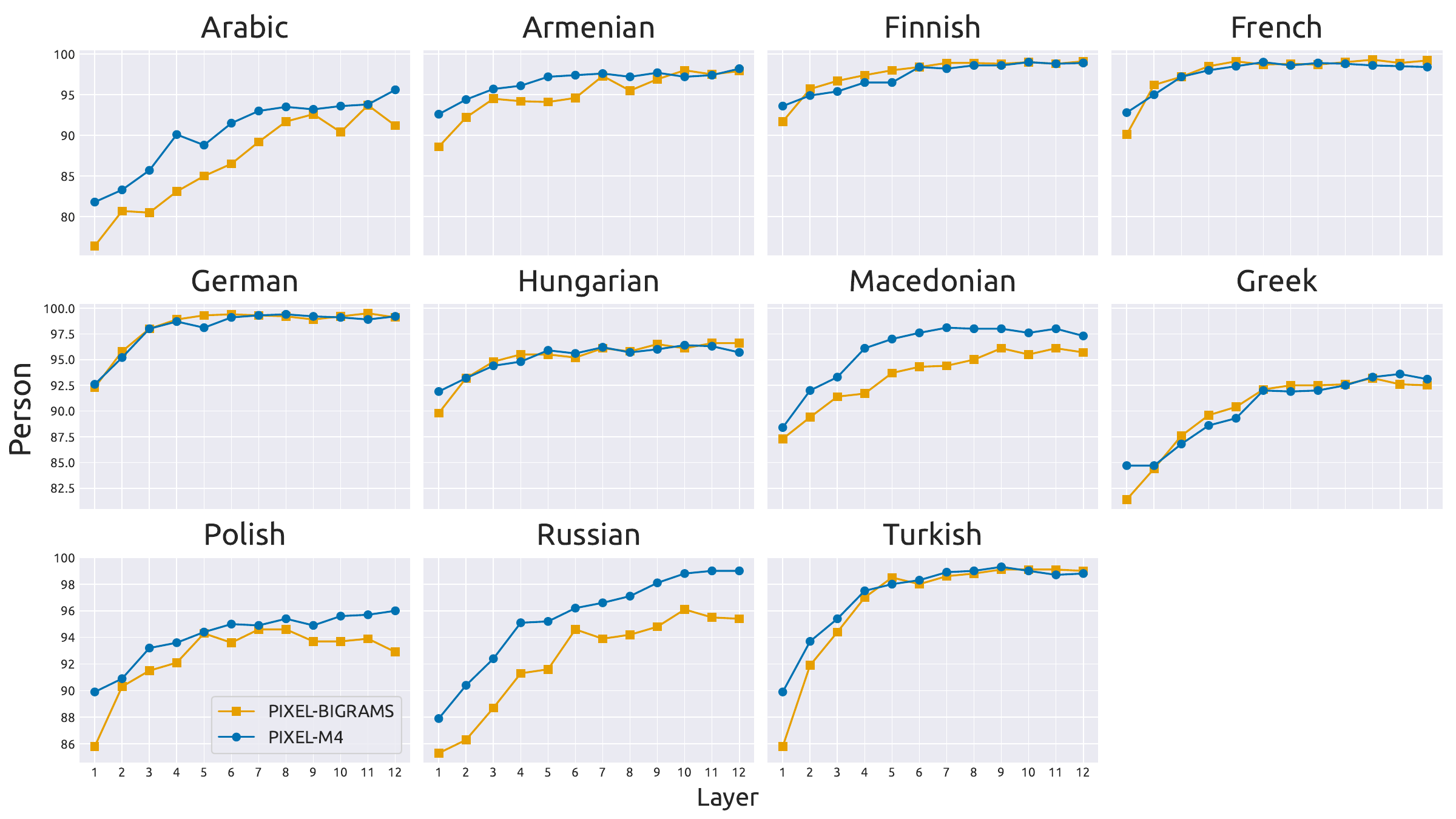}
  \caption{Word-level probing analysis on \textsc{linspector} for the Person task. Each subplot shows a different language; in each, the y-axis represents model accuracies and the x-axis represents layer number of the hidden representations.}
  \label{fig:linspector-appendix-person}
\end{figure*}

\begin{figure*}[htbp]
  \centering
  \includegraphics[width=\textwidth]{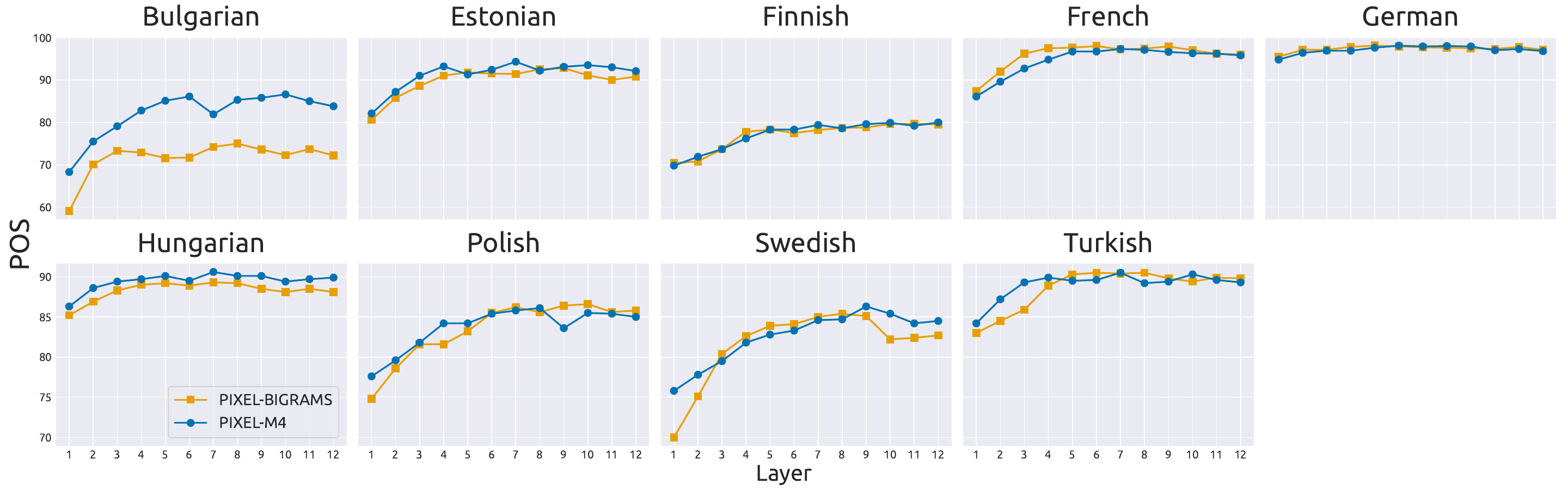}
  \caption{Word-level probing analysis on \textsc{linspector} for the Pos task. Each subplot shows a different language; in each, the y-axis represents model accuracies and the x-axis represents layer number of the hidden representations.}
  \label{fig:linspector-appendix-pos}
\end{figure*}

\begin{figure*}[htbp]
  \centering
  \includegraphics[width=\textwidth]{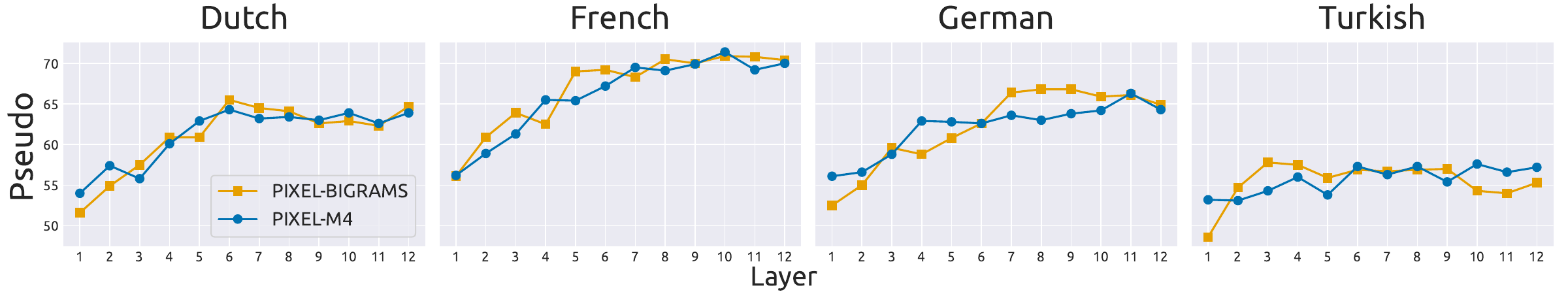}
  \caption{Word-level probing analysis on \textsc{linspector} for the Pseudo task. Each subplot shows a different language; in each, the y-axis represents model accuracies and the x-axis represents layer number of the hidden representations.}
  \label{fig:linspector-appendix-pseudo}
\end{figure*}

\begin{figure*}[htbp]
  \centering
  \includegraphics[width=\textwidth]{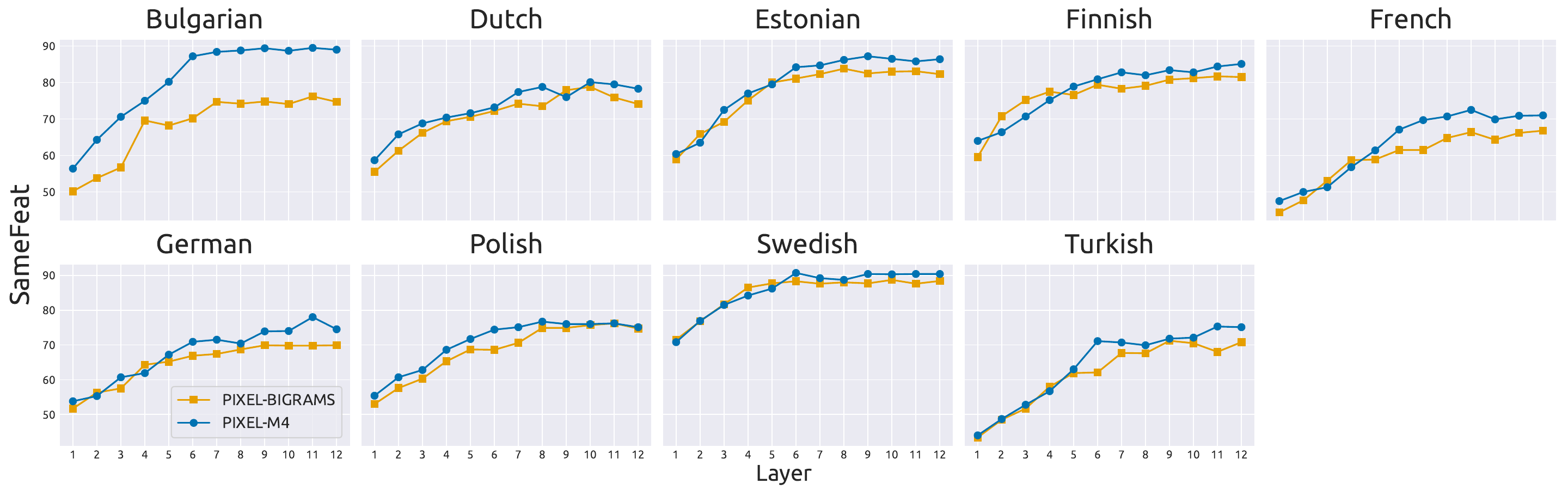}
  \caption{Word-level probing analysis on \textsc{linspector} for the SameFeat task. Each subplot shows a different language; in each, the y-axis represents model accuracies and the x-axis represents layer number of the hidden representations.}
  \label{fig:linspector-appendix-samefeat}
\end{figure*}

\begin{figure*}[htbp]
  \centering
  \includegraphics[width=\textwidth]{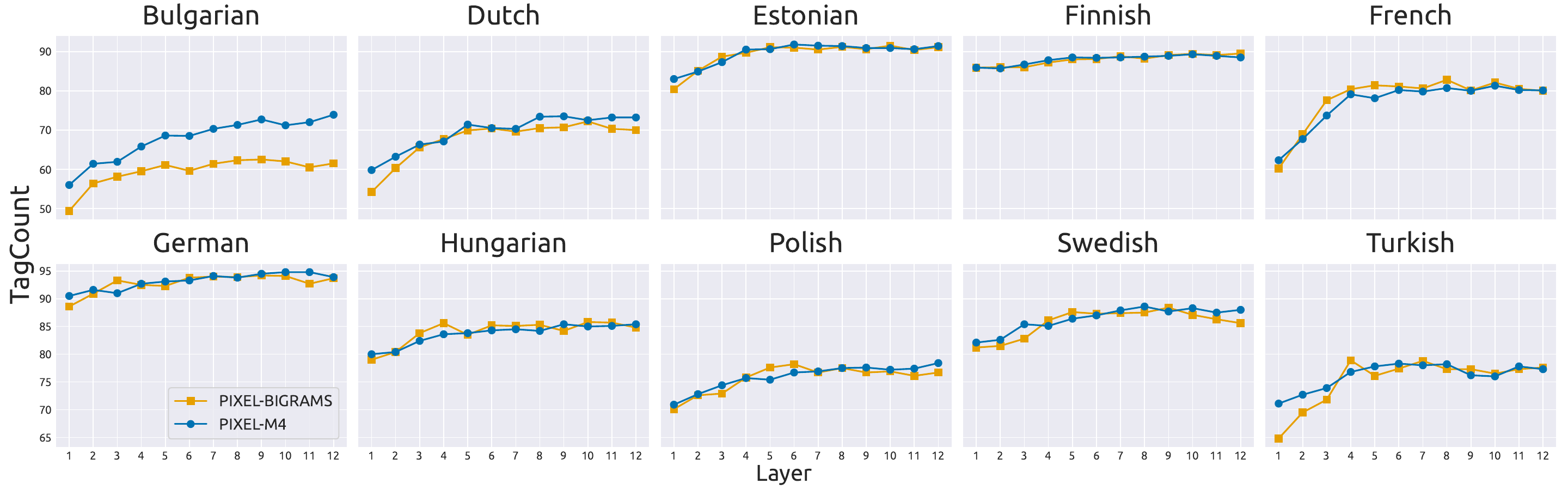}
  \caption{Word-level probing analysis on \textsc{linspector} for the TagCount task. Each subplot shows a different language; in each, the y-axis represents model accuracies and the x-axis represents layer number of the hidden representations.}
  \label{fig:linspector-appendix-tagcount}
\end{figure*}

\begin{figure*}[htbp]
  \centering
  \includegraphics[width=\textwidth]{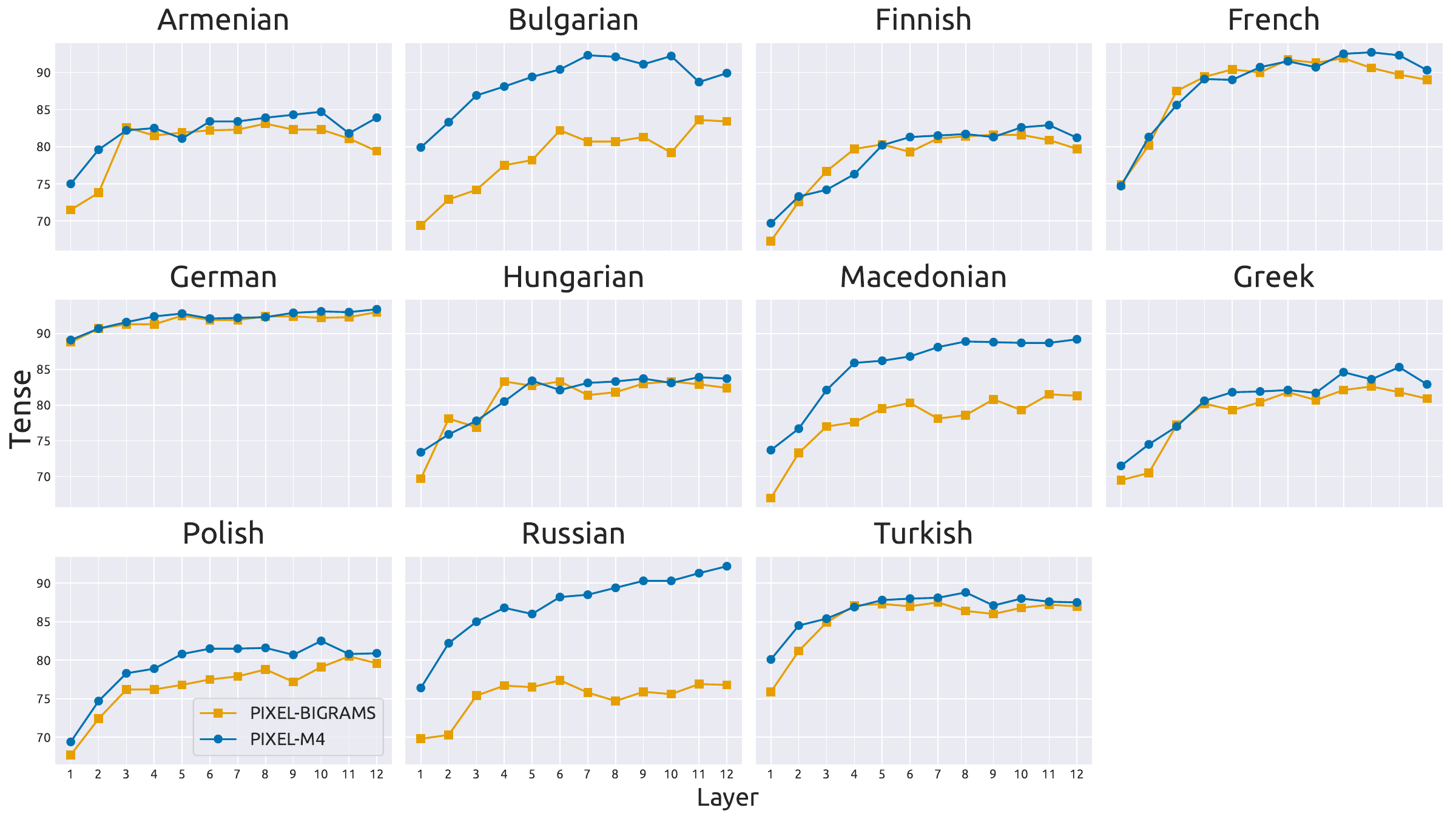}
  \caption{Word-level probing analysis on \textsc{linspector} for the Tense task. Each subplot shows a different language; in each, the y-axis represents model accuracies and the x-axis represents layer number of the hidden representations.}
  \label{fig:linspector-appendix-tense}
\end{figure*}

\begin{figure*}[htbp]
  \centering
  \includegraphics[width=\textwidth]{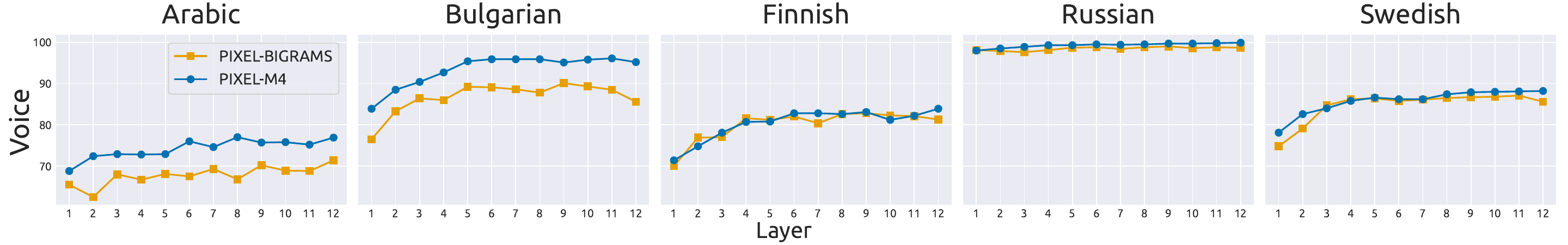}
  \caption{Word-level probing analysis on \textsc{linspector} for the Voice task. Each subplot shows a different language; in each, the y-axis represents model accuracies and the x-axis represents layer number of the hidden representations.}
  \label{fig:linspector-appendix-voice}
\end{figure*}

\begin{figure*}[htbp]
  \centering
  \includegraphics[width=0.8\textwidth]{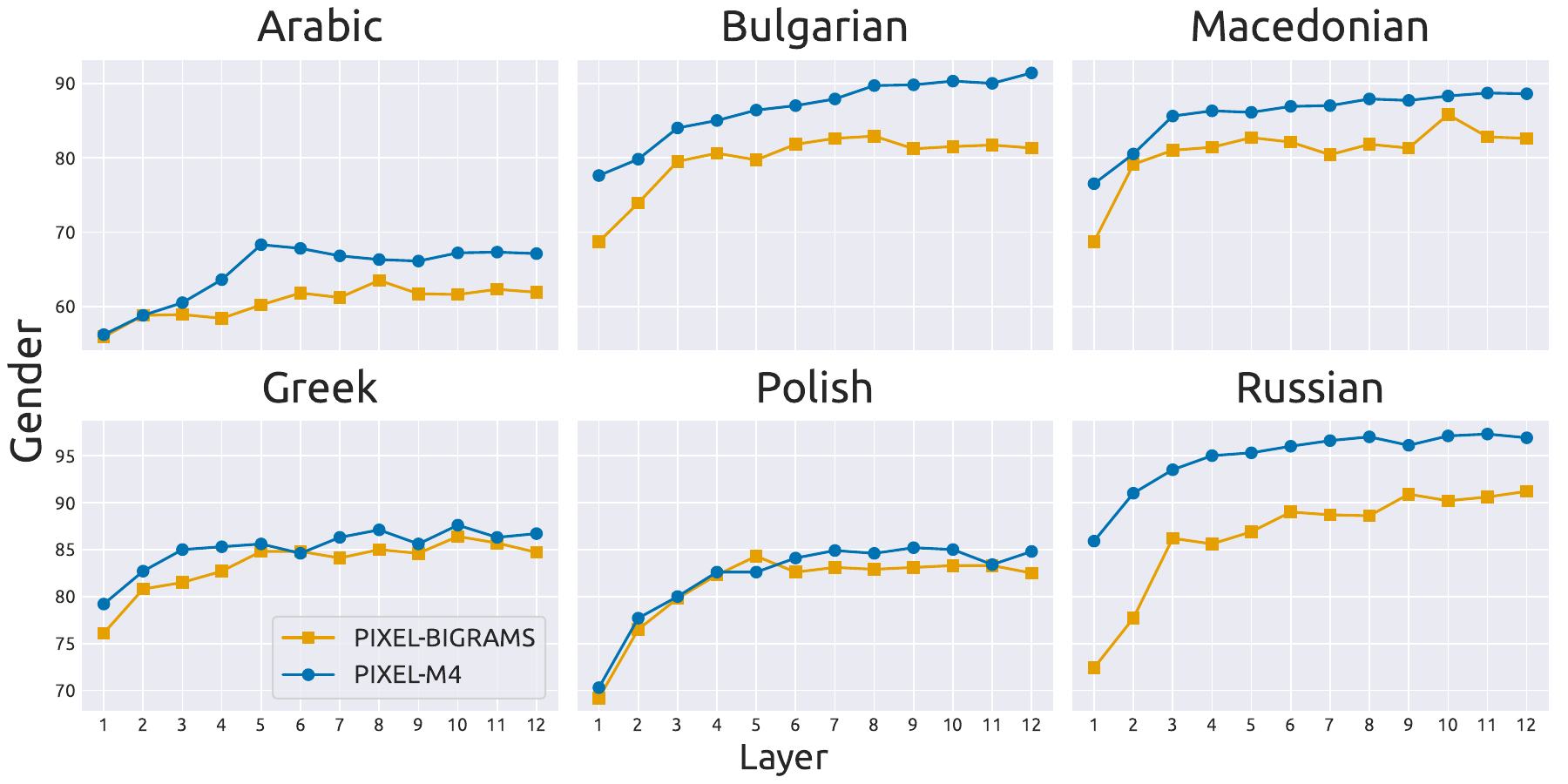}
  \caption{Word-level probing analysis on \textsc{linspector} for the Gender task. Each subplot shows a different language; in each, the y-axis represents model accuracies and the x-axis represents layer number of the hidden representations.}
  \label{fig:linspector-appendix-gender}
\end{figure*}

\begin{figure*}[htbp]
  \centering
  \begin{minipage}{\columnwidth}
    \centering
    \includegraphics[width=0.75\linewidth]
      {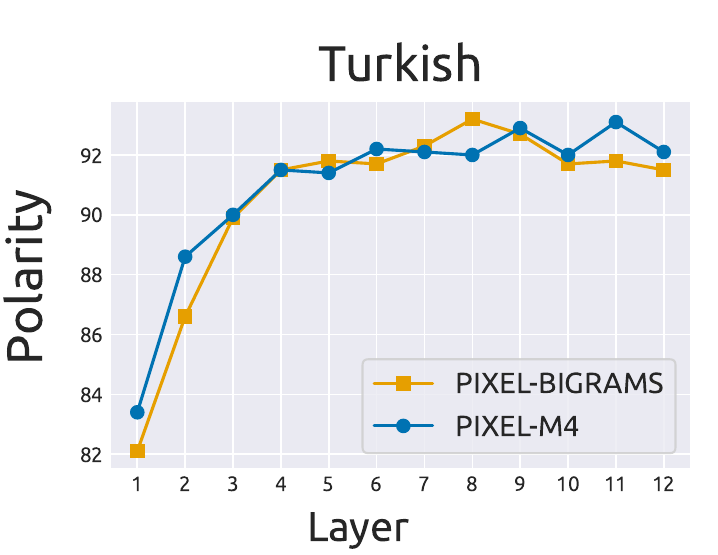}
    \caption{Word‑level probing analysis on \textsc{linspector} for the Polarity task. 
      Each subplot shows a different language; in each, the y‑axis represents model
      accuracies and the x‑axis represents layer number of the hidden representations.}
    \label{fig:linspector-appendix-polarity}
  \end{minipage}
\end{figure*}

\begin{figure*}[htbp]
  \centering
  \begin{minipage}{\columnwidth}
    \centering
    \includegraphics[width=\linewidth]
      {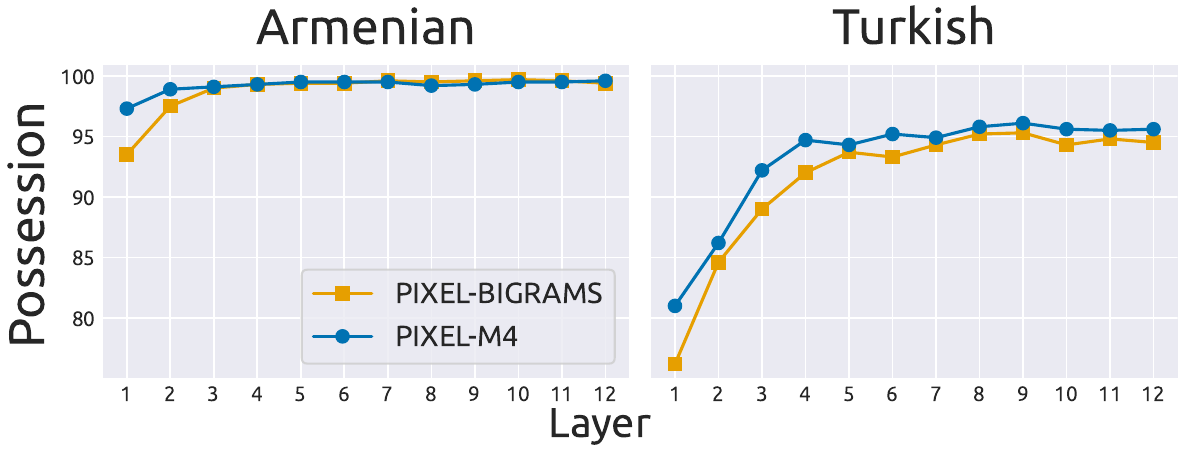}
    \caption{Word‑level probing analysis on \textsc{linspector} for the Possession task. 
      Each subplot shows a different language; in each, the y‑axis represents model
      accuracies and the x‑axis represents layer number of the hidden representations.}
    \label{fig:linspector-appendix-possession}
  \end{minipage}
\end{figure*}

In this appendix section, we share the results for the rest of the word-level probing analyses on \textsc{linspector}~\citep{csahin-etal-2020-linspector}.
We analyze our model on fifteen languages—Arabic, Armenian, Bulgarian, Dutch, Estonian, Finnish, French, German, Greek, Hungarian, Macedonian, Polish, Russian, Swedish, and Turkish—across fourteen linguistic probing tasks: \textit{Case Marking} (Fig.~\ref{fig:linspector-appendix-case}), \textit{Gender} (Fig.~\ref{fig:linspector-appendix-gender}), \textit{Mood} (Fig.~\ref{fig:linspector-appendix-mood}), \textit{Number} (Fig.~\ref{fig:linspector-appendix-number}), \textit{OddFeat} (Fig.~\ref{fig:linspector-appendix-oddfeat}), \textit{Person} (Fig.~\ref{fig:linspector-appendix-person}), \textit{Polarity} (Fig.~\ref{fig:linspector-appendix-polarity}), \textit{POS} (Fig.~\ref{fig:linspector-appendix-pos}), \textit{Possession} (Fig.~\ref{fig:linspector-appendix-possession}), \textit{Pseudo} (Fig.~\ref{fig:linspector-appendix-pseudo}), \textit{SameFeat} (Fig.~\ref{fig:linspector-appendix-samefeat}), \textit{TagCount} (Fig.~\ref{fig:linspector-appendix-tagcount}), \textit{Tense} (Fig.~\ref{fig:linspector-appendix-tense}), and \textit{Voice} (Fig.~\ref{fig:linspector-appendix-voice}).

These analyses provide further support for the findings reported in \S\ref{sec:analysis}.
Throughout the entire network, \mpixel captures more robust linguistic features than \pixelbigrams on all tasks for the Cyrillic script languages, Bulgarian, Macedonian and Russian.
This is again expected since \mpixel has seen a similar language, e.g. Ukrainian, during pretraining.
Similarly, our observations are the same for the languages in unseen scripts, Arabic, Armenian and Greek, showcasing the improved cross-lingual transfer learning capabilities of \mpixel.
Furthermore, on Latin script languages, both models achieve similar overall performances across the layers.
Nonetheless, on some tasks, \mpixel captures better linguistic features for Latin languages with diacritics (e.g. Turkish, Swedish).
Additionally, on more complex tasks such as \textit{OddFeat} and \textit{SameFeat}, \mpixel outperforms \pixelbigrams on Latin script languages like German and Hungarian, where the two models perform similarly on the other tasks.

\subsection{Use of AI Assistants}
Within this work, we used AI assistants only to generate code for producing the plots.

\end{document}